\documentclass[sigconf]{acmart}
\usepackage{amsmath}

\usepackage{amssymb}
\usepackage{amsfonts}
\usepackage{algorithm}
\usepackage{algpseudocode}
\usepackage{graphicx}
\usepackage{textcomp}
\usepackage{xcolor}
\usepackage{multirow}
\usepackage{subfig}

\AtBeginDocument{%
  }

\setcopyright{acmcopyright}
\copyrightyear{2024}
\acmYear{2024}
\setcopyright{rightsretained}
\acmConference[CF '24]{21st ACM International Conference on Computing
Frontiers}{May 7--9, 2024}{Ischia, Italy}
\acmBooktitle{21st ACM International Conference on Computing Frontiers (CF
'24), May 7--9, 2024, Ischia, Italy}
\acmDOI{10.1145/3649153.3649212}
\acmISBN{979-8-4007-0597-7/24/05}

\acmSubmissionID{123-A56-BU3}

\begin{document}

\title{TabConv: Low-Computation CNN Inference via Table Lookups}


\author{Neelesh Gupta}
\authornote{These authors contributed equally.}
\affiliation{%
  \institution{University of Southern California}
  \city{Los Angeles}
    \state{California}
  \country{USA}
  \postcode{90089}
}
\email{neeleshg@usc.edu}

\author{Narayanan Kannan}
\authornotemark[1]
\affiliation{
  \institution{University of California, Los Angeles}
  \city{Los Angeles}
  \state{California}
  \country{USA}
  \postcode{90095}}
\email{nkann19@ucla.edu}

\author{Pengmiao Zhang}
\authornotemark[1]
\authornote{Corresponding author.}
\affiliation{
  \institution{University of Southern California}
  \city{Los Angeles}
  \state{California}
  \country{USA}
  \postcode{90089}
}
\email{pengmiao@usc.edu}

\author{Viktor Prasanna}
\affiliation{%
  \institution{University of Southern California}
  \city{Los Angeles}
  \state{California}
  \country{USA}
}
\email{prasanna@usc.edu}

\renewcommand{\shortauthors}{Gupta et al.}

\newcommand{\ourwork}{TabConv}

\begin{abstract}
Convolutional Neural Networks (CNNs) have demonstrated remarkable ability throughout the field of computer vision. 
However, CNN inference requires a large number of arithmetic operations, making them expensive to deploy in hardware.
Current approaches alleviate this issue by developing hardware-supported, algorithmic processes to simplify spatial convolution functions. 
However, these methods still heavily rely on matrix multiplication, leading to significant computational overhead.
To bridge the gap between hardware, algorithmic acceleration, and approximate matrix multiplication, we propose \textit{TabConv}, a novel, table-based approximation for convolution to significantly reduce arithmetic operations during inference.
Additionally, we introduce a priority masking technique based on cosine similarity to select layers for table-based approximation, thereby maintaining the model performance.
We evaluate our approach on popular CNNs: ResNet-18, ResNet-34, and NetworkInNetwork (NIN).
TabConv preserves over 93\% of the original model's performance while reducing arithmetic operations by 36.5\%, 25.8\%, and 99.4\% for ResNet-18 on CIFAR-10, CIFAR-100, and MNIST, respectively, 35.6\% and 99.3\% for ResNet-34 on CIFAR-10 and MNIST, and 98.9\% for NIN on MNIST, achieving low-computation inference.
\end{abstract}

\ccsdesc[500]{Computing methodologies~Neural networks}
\ccsdesc[500]{Computing methodologies~Computer vision}

\keywords{convolutional neural network, table lookup, product quantization}


\maketitle

\section{Introduction}
Convolutional Neural Networks (CNNs) are ubiquitous in the field of computer vision. They are widely adopted as solutions to tasks in image classification~\cite{russakovsky2015imagenet, touvron2019fixing}, object detection~\cite{girshick2015fast}, semantic segmentation~\cite{long2015fully, yuan2020object}, computer systems~\cite{zhang2022fine}, et cetera. 
{\color{black}
This utility has fueled concerted efforts to enhance CNN inference, yielding new models that surpass benchmarks.
Unfortunately, these improvements result in a significant increase in computational cost.}

The continued expansion of CNN model complexity has caused an explosion in both the number of operations and amount of time required to process data and make predictions \cite{he2016deep, simonyan2015deep, krizhevsky2012imagenet}. Moreover, this issue has all but rendered the traditional, multi-core CPU obsolete as a deep learning platform, thus mandating the usage of other dedicated hardware devices such as GPUs \cite{kim2017performance} and FPGAs \cite{putnam2015reconfigurable} to power model operations. However, they too face their own unique set of challenges in energy consumption and a lack of standardization, respectively. Given these issues, much research into mitigating CNN computational costs focuses not on hardware, but on optimizing functions of the model itself.


Many of these efforts can be summarized as attempts to simplify the convolution function through mapping the process to an already well-optimized operation, such as matrix multiplication (MM) \cite{chetlur2014cudnn, chellapilla2006high}. Though they achieve a definite reduction in complexity, these methods leave ample room for improvement. High computational costs remain a bottleneck for inference latency, even when limited to refined processes. To combat this, techniques in approximating MM have been incorporated into Deep Neural Network (DNN) workflows, including those that simplify computation~\cite{drineas2001fast} and those that replace it entirely \cite{blalock2021multiplying, elhoushi2021deepshift, zhang2024attention}. 
A promising approach involves the use of Product Quantization (PQ) to map MMs to table lookups~\cite{jegou2010product}. However, existing works~\cite{blalock2021multiplying,yu2018product} map only the MM of the last linear layer in a DNN to a table and suffer a large dropoff in accuracy when attempting more.


Motivated by shortcomings in both hardware and algorithm-based acceleration as well as the promise of table-based computation,
we introduce \textit{TabConv}, a novel low-computation CNN approximation based on table lookups that significantly reduces MMs in a CNN model while maintaining its performance.
To achieve this, we follow a three-step process that integrates aspects of prior works with our own contributions: 1) Taking a trained CNN, converting all instances of convolution into MMs, 2) Mapping these MMs to table lookups based on product quantization, 3) Employing a novel priority masking approach to determine which layers should retain their exact computations rather than table-based approximations. 



We solve two key problems during the development of TabConv-based CNN approximations.
1) The issue of mapping entire neural network layers with various operations--We introduce novel table approximation solutions for convolution, linear operations, batch normalization, and activation functions. 
2) The compounding increase in error by layer once mapped to a table-based approximation--We combat this issue using a novel priority masking method that identifies which layers are best left in neural network form. 

We summarize our main contributions below:
\begin{itemize}
    \item We propose \ourwork, a novel low-computation CNN approximation through table lookups, which significantly reduces the arithmetic operations required during inference while maintaining the CNN model performance.
    \item We design tabular primitives for operations in CNNs models, including convolution, batch normalization, linear operations, and activation functions.
    \item We propose a novel priority masking strategy that selectively converts certain layers in a model to table lookups while retaining other layers as exact calculations, balancing the trade-off between computation and accuracy.
    \item We evaluate~\ourwork~on popular CNNs ResNet-18, ResNet-34, and NetworkInNetwork (NIN). Results show that while preserving higher than 93\% of the original model performance,~\ourwork~reduces 36.5\%, 25.8\%, and 99.4\% of arithmetic operations for ResNet-18 on CIFAR-10, CIFAR-100, and MNIST, reduces 35.6\% and 99.3\% of arithmetic operations for ResNet-34 on CIFAR-10 and MNIST, and reduces 98.9\% of arithmetic operations for NIN on MNIST.

\end{itemize}

\section{Background}

\subsection{Product Quantization}
\label{sec:background-pq}
Our approach is built upon the Product Quantization (PQ) algorithm \cite{jegou2010product}, notable for its use in accelerating and approximating vector inner products through quantization and precomputation. In general, given an arbitrary vector $\mathbf{a}\in\mathbb{R}^D$ drawn from a training set $\Tilde{\mathbf{A}}\in\mathbb{R}^{N\times D}$ with $N$ samples, and a fixed weight vector $\mathbf{b}\in\mathbb{R}^D$, PQ generates a quantized approximation $\hat{\mathbf{a}}$ of $\mathbf{a}$ such that $\hat{\mathbf{a}}^\top\mathbf{b}\approx\mathbf{a}^\top\mathbf{b}$. In order to carry out this process, $D$-dimensional $\mathbf{a}$ is split into $S$ disjoint, $V$-dimensional subspaces, within each of which $K$ quantized subvectors are learned as prototypes. Note that since $\hat{\mathbf{a}}$ is quantized and $\mathbf{b}$ is fixed, their corresponding inner product is easily precomputed and reused in a query. Figure 1 provides an overview of the PQ process, while a detailed description follows below. 

\begin{figure}[h]
    \centering
    \subfloat[During PQ training, prototypes within each subspace are learned and their dot products with weight vector $\mathbf{b}$ are precomputed for storage in a table.]
    {\includegraphics[width=\linewidth]{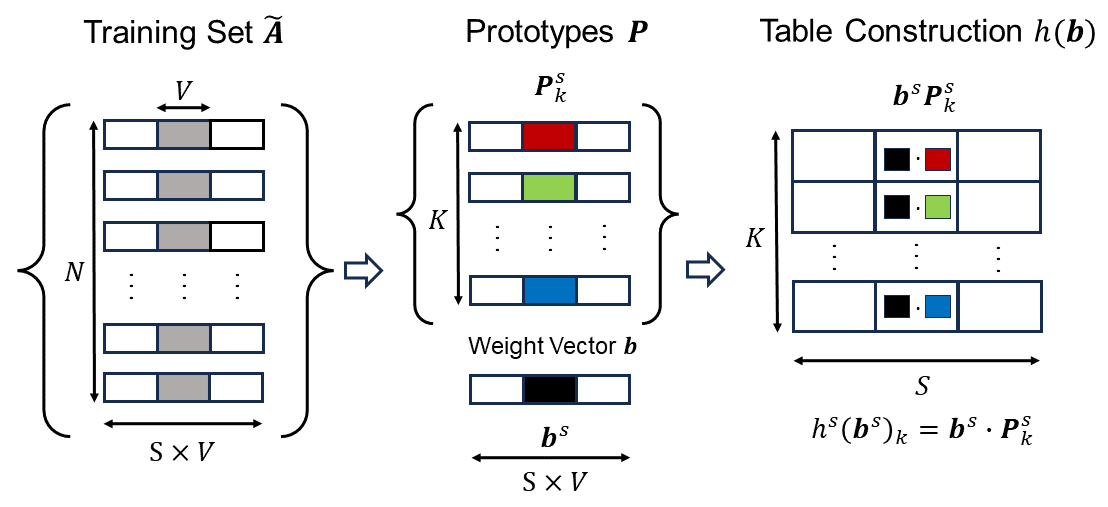}}\label{fig:pq_train}
    \newline
    \subfloat[During PQ query, the query vector is encoded to find the indices of its nearest prototypes, whose dot products are then looked up in the table and subsequently aggregated to yield the final result.]
    {\includegraphics[width=\linewidth]{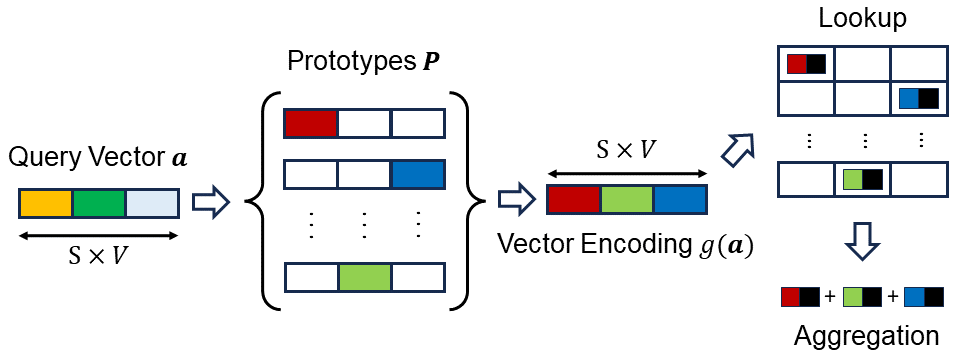}\label{fig:pq_query}}
    \caption{Training and query of product quantization.}
    \label{fig:pq}
\end{figure}

\subsubsection{Training}
The PQ training process includes both a prototype learning phase and table construction phase.

\noindent\textbf{Prototype Learning ($p$)}: Consider $\Tilde{\mathbf{A}}^\mathbf{s} \in\mathbb{R}^{N\times V}$, the vectors of the $s$-th subspace of $\Tilde{\mathbf{A}}$. 
{
\color{black}
In the Prototype Learning phase, $K$ prototypes, $\mathbf{P}_k^s$, where $k$ is the index of the prototypes within the $s$-th subspace, of $\Tilde{\mathbf{A}}^s$ are learned by minimizing the distance between the vectors of $\Tilde{\mathbf{A}}^s$ and their nearest corresponding prototype $\mathbf{P}_k^s$.
}
The process is formulated in Equation \ref{eq:pq-1} below. 
\begin{equation}
\label{eq:pq-1}
    p^c(\Tilde{\mathbf{A}}) \triangleq \underset{P}{\arg \min } \sum_s \sum_i\left\|\Tilde{\mathbf{A}}_i^s-\mathbf{P}_k^s\right\|^2
\end{equation}

\noindent\textbf{Table Construction ($h$)}: 
{\color{black}
Next, we construct a table with entries consisting of inner products between prototypes $\mathbf{P}_k^s$ and the weight vector $\mathbf{b}^s$ where $s$ signifies the vector belonging to the $s$-th subspace. }
The function $h^{s}(\mathbf{b})_k$ describes the $sk$-th entry of the table. 
\begin{equation}
\label{eq:pq-2}
    h^{s}(\mathbf{b})_k \triangleq {\mathbf{b}^s}^\top \cdot \mathbf{P}_{k}^{s}
\end{equation}

\subsubsection{Query}
The query process eliminates the need for multiplication operations in the inner product calculation by encoding the query vector to its nearest prototype, looking up its corresponding table entries, and aggregating to yield the final result.

\noindent\textbf{Vector Encoding ($g$)}: For arbitrary query vector $\mathbf{a}$, $g^s(\mathbf{a})$ locates its closest prototype $\mathbf{P}_k^s$ in each subspace $s$ by finding the index $k$ with minimal distance to $\mathbf{a}^s$. The function, as formulated below, outputs a set of indices representing the encoded vector of $\mathbf{a}$.
\begin{equation}
\label{eq:pq-3}
    g^s\left(\mathbf{a}\right) \triangleq \underset{k}{\arg \min }\left\|\mathbf{a}^s-\mathbf{P}_k^s\right\|^2
\end{equation}

\noindent\textbf{Lookup and Aggregation ($f$)}: After using the encoded indices to lookup the precomputed values, the corresponding entries by subspace are aggregated through the following function $f(\cdot, \cdot)$, yielding a final approximation for $\mathbf{a}^\top\mathbf{b}$.
\begin{equation}
\label{eq:pq-4}
    f\left(\mathbf{a}, \mathbf{b}\right) = \sum_s h^{s}\left(\mathbf{b}\right)_k, k = g^{s}\left(\mathbf{a}\right)
\end{equation}

Thus, the actual dot product operation in $\mathbf{a}^\top\mathbf{b}$ is avoided by approximating the result through table lookups. We use locality sensitive hashing \cite{blalock2021multiplying} for encoding and parallel summation for aggregation, resulting in significantly lower complexity than the dot product, especially for large dimensional vectors.


\subsection{Im2col Convolution}
\label{sec:background-im2col}
The im2col method \cite{chetlur2014cudnn, chellapilla2006high} transforms the convolution function into a general instance of matrix multiplication (MM) by converting both the input image and kernel into patch matrices. An outline of this process is depicted in Figure 2. For simplicity's sake, we consider the case when stride is $1$ and padding is $0$.
\begin{figure}[h]
    \centering
    {\includegraphics[width=\linewidth]{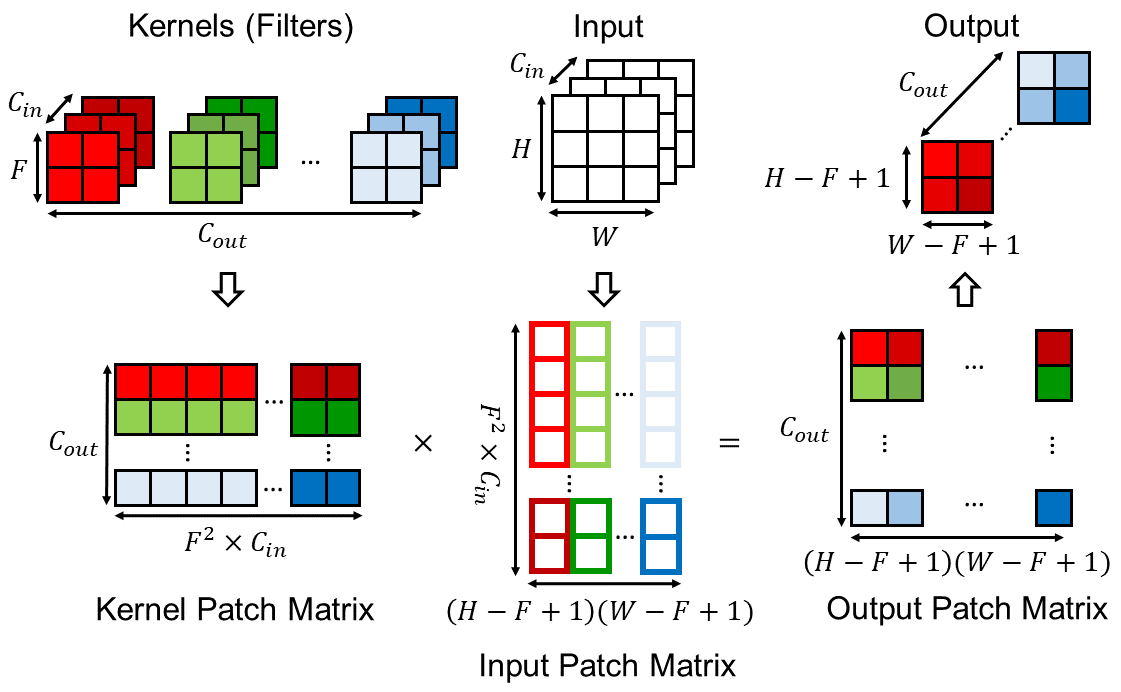}}\label{fig:im2col} 
    \caption{Outline of the im2col Method.}
    \label{fig:im2col_outline}
\end{figure} 

\noindent\textbf{Input Patch Matrix.} Given an input image with $C_{in}$ channels of size $H\times W$ and $C_{out}$ kernels (filters) of size $F\times F$, the input patch matrix is formed by taking kernel-sized patches of each input channel and flattening them into column vectors which are then concatenated to form a new matrix of size ${{F^2}C_{in}}\times{(H-F+1)(W-F+1)}$.

\noindent\textbf{Kernel Patch Matrix.} The kernel patch matrix is formed by reshaping each channel of each kernel into row vectors which are then concatenated similarly as above. The result is of size $C_{out}\times {{F^2}{C_{in}}}$. 

We apply im2col as an intermediate step to transform the convolution function into an MM operation. Following this conversion, we replace subsequent MMs with approximate table lookups.


\section{Related Work}
\subsection{CNN Acceleration}




There exists a rich body of literature on techniques to accelerate CNNs.
Of particular note are those utilizing algorithmic processes, model compression methods, and hardware implementation.
Many algorithm-based methods implement a mapping of the convolution function to an instance of matrix multiplication (MM), including the following processes: the im2col transformation \cite{chetlur2014cudnn, rohwedder2021pooling, chellapilla2006high}, kn2row transformation \cite{vasudevan2017parallel}, Winograd Minimal Filtering Method \cite{lavin2016fast}, Toeplitz matrix conversion \cite{chen2020survey}, and Fast Fourier Transform convolutions \cite{chi2020fast, ko2017design}. 
Model compression techniques like pruning \cite{liang2021pruning, radu2019performance}, quantization \cite{han2015deep, ruospo2021investigating}, and knowledge distillation~\cite{hinton2015distilling, gupta2023packd} reduce redundancy and model size to help performance. 
Acceleration through hardware entails the usage of GPUs with high performance computing capability \cite{nurvitadhi2017can, rovder2019optimising}, and energy efficient FPGAs offering parallel acceleration tailored for CNNs \cite{abdelouahab2018accelerating, sterpone2023cnnoriented}. 
While these methods are successful in accelerating computation, they heavily feature MM. Eliminating these costly operations offers a new opportunity for advancement in CNN acceleration. In this paper, we introduce a novel approach that maps CNN to a series of fast tabular lookups.

\subsection{Approximate Matrix Multiplication}





Techniques in approximate matrix multiplication use algorithmic methods to simplify computation. For example, sampling input matrices \cite{drineas2001fast}, finding sketches of matrices \cite{ye2016frequent, mroueh2017co}, and random projection to lower dimensional subspaces \cite{magen2011low, cohen2015optimal} all attempt to reduce the number of rows or columns being operated on. Other approaches go further by replacing MM outright through techniques in hashing, averaging, logarithm computation, and designing distributed algorithms \cite{blalock2021multiplying, francis2022practical, kim2021effects, li2021efficient}. Notably, the Product Quanitzation (PQ) algorithm \cite{jegou2010product} is used to convert traditional instances of MM into a series of tabular lookups \cite{blalock2021multiplying}. 
{\color{black}Our work presents a comprehensive design methodology and acceleration strategy that both maintains and accelerates performance when applying PQ to CNNs.}

\section{Approach}

\subsection{Problem Definition}

Our objective is to refine CNN inference by employing table lookups to approximate its computations. 
{\color{black}
Let $\mathcal{M}$ be a CNN model characterized by its parameters $\boldsymbol{\theta}$ where $\mathcal{M}(\mathbf{x}; \boldsymbol{\theta})$ represents the model's output given input $\mathbf{x}$.
Our objective is to construct a table-based approximation $\mathcal{T}$ with parameters $\boldsymbol{\phi}$ such that $\mathcal{T}(\mathbf{x}; \boldsymbol{\phi})$ closely approximates the output of $\mathcal{M}$.
}
We formalize this as follows:
\begin{equation}
\begin{aligned}
    \min_{\boldsymbol{\phi}} \left[ \frac{1}{N} \sum_{i=1}^{N} \left\| \mathcal{M}(\mathbf{x}_i; \boldsymbol{\theta}) - \mathcal{T}(\mathbf{x}_i; \boldsymbol{\phi}) \right\|^2 \right] \\
\end{aligned}
\end{equation}
We aim to minimize the discrepancy between the outputs of the conventional and the table-based models, ensuring that the latter is able to reduce arithmetic operations while maintaining accuracy.

\subsection{Overview of TabConv}

\begin{figure}[ht]
    \centering
    {\includegraphics[width=\linewidth]{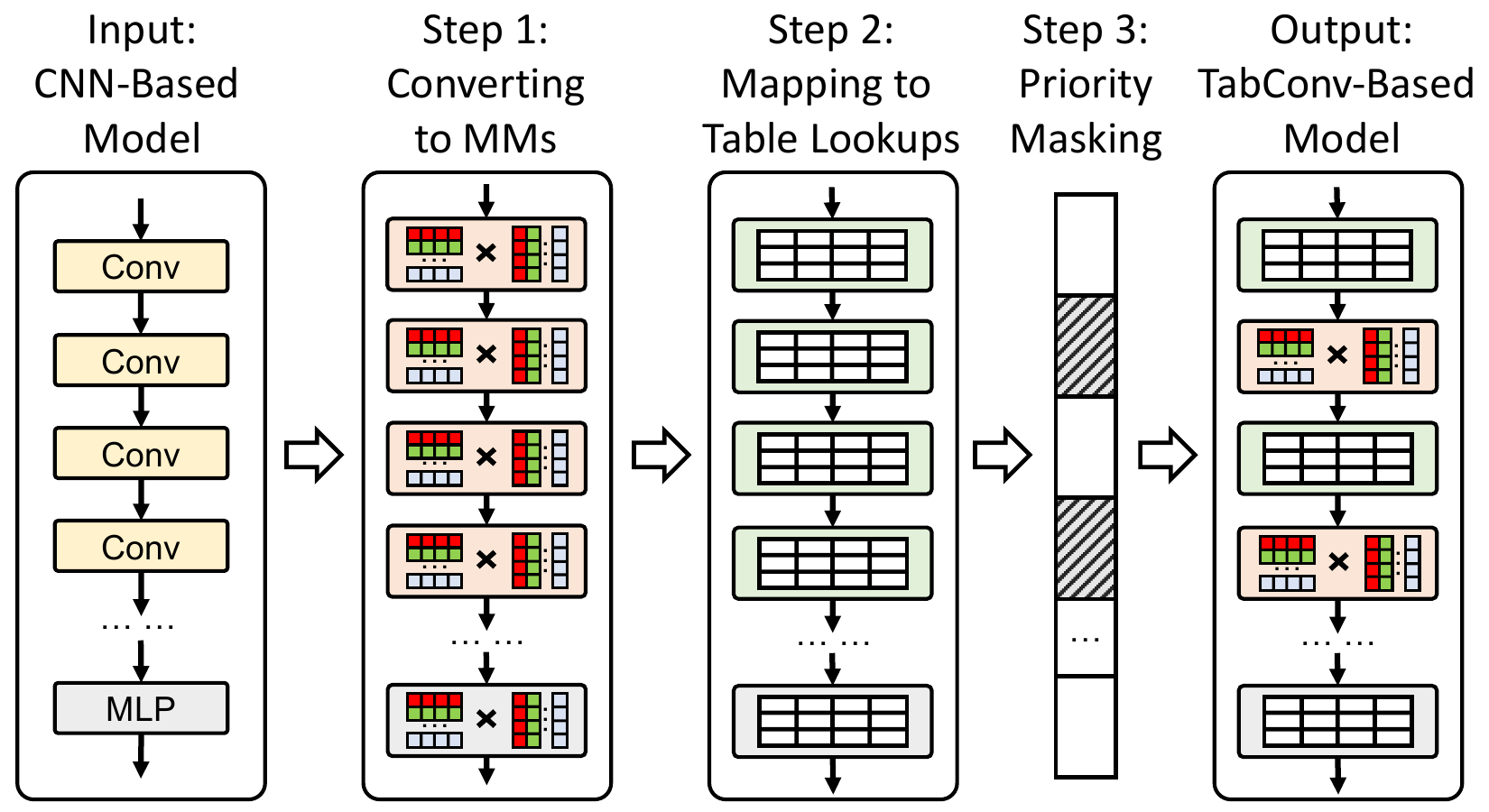}}
    \caption{Workflow of converting a CNN-based model to the proposed TabConv-based model.}
    \label{fig:overview}
\end{figure}

For a given CNN-based model, we generate its approximation, a TabConv-based model, through a three-step strategy, as depicted in Figure~\ref{fig:overview}, including: 1) Converting convolution operations within a model into matrix multiplications (MM) format; 2) Mapping the resulting MMs to table lookups; 3) Employing a novel priority masking strategy to strike a balance between accuracy and computation. Next, we introduce the detailed workflow.

\noindent\textbf{Input: CNN-based model.} 
The input to the TabConv process is a CNN-based model that has already been trained for a specific task. This model likely consists of convolutional layers, batch normalization, linear layers, and activation functions, depending on the architecture and task it was designed for. 

\noindent\textbf{Step 1: Converting to MMs.} 
In the initial step, the key operations within the original CNN-based model are transformed into the format of MMs. For convolution operations, we use im2col \cite{chetlur2014cudnn} as described in Section~\ref{sec:background-im2col}. For batch normalization operations, we fold the operation into the im2col MMs by merging the normalized weights into the convolution weights (Section~\ref{sec:approach-bn}). This conversion facilitates the subsequent steps by providing a more structured and efficient representation of the operations involved in the model.

\noindent\textbf{Step 2: Mapping to table lookups.} 
The resulting MMs are then mapped to table lookups based on product quantization (Section~\ref{sec:background-pq}). We quantize the input data of each CNN and linear layer into a fixed size of prototypes (Section~\ref{sec:cnn-pq}), precompute the dot product of the quantized vectors and the layer weights, and store the results in a table. In inference, we lookup the layer results from the closest prototypes to the input vectors, avoiding any MM operations.

\noindent\textbf{Step 3: Priority masking.} 
In deep CNNs, increasing the number of layers mapped to table lookups can lead to degraded approximation performance and accuracy, as deeper layers are likely to compound existing input errors.
To address this, we introduce a novel priority masking method, employing a "similarity drop" metric to quantify differences between table and original CNN layers (Section~\ref{sec:masking}).


\noindent\textbf{Output: TabConv-based model.} 
The output of this process is the TabConv-based model, an approximation of the original CNN-based model. This TabConv-based model closely approximates the predictive performance of the original model while significantly reduces the number of arithmetic operations involved in inference.

\subsection{Mapping CNN to Table Lookups}
\label{sec:cnn-pq}


We map operations in CNN to table lookups based on im2col and product quantization. In the following, we describe the process of constructing table 
CNN operations and how the table-based inference eliminates MM operations.

\subsubsection{Table Construction for Convolution}
\begin{figure}[ht]
    \centering
    {\includegraphics[width=\linewidth]{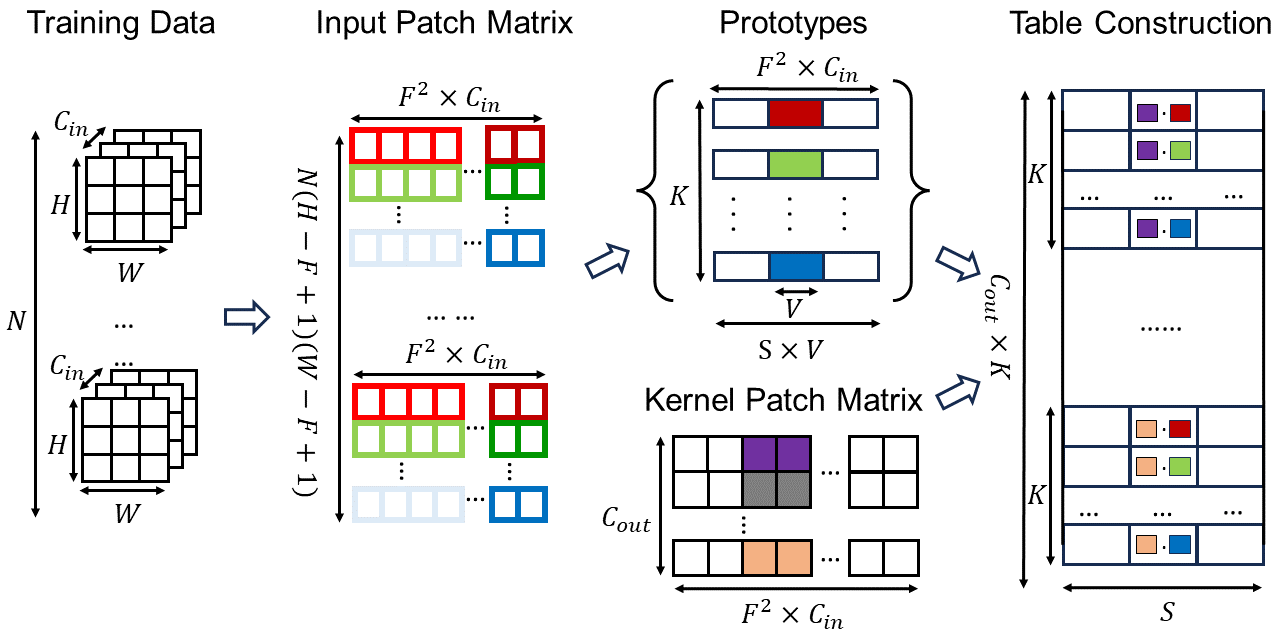}}
    \caption{Table construction for convolutional operation.}
    \label{fig:conv_pq_train}
\end{figure}

Figure~\ref{fig:conv_pq_train} illustrates the table construction for precomputed convolutional results.
The im2col algorithm reshapes convolutional layer inputs from $C_{in} \times H\times W$ to $HW \times F^2C_{in}$ (assuming with paddings). 
This process creates a 2D matrix of dimensions $NHW\times F^2C_{in}$ by combining $N$ data points from the training dataset. 
The $F^2C_{in}$ dimension is divided into $S$ subspaces, each learning $K$ prototypes through methods like unsupervised clustering or locality-sensitive hashing~\cite{blalock2021multiplying}.
After learning the $K$ prototypes for each subspace, dot products are computed between these prototypes and the corresponding subspace of the kernel patch matrix, reshaped from $C_{out}\times F^2C_{in}$ kernel weights.
This results in $C_{out}$ sub-tables, one for each output channel, storing dot products between prototypes and channel weights with each entry representing the product for a subspace.

\begin{figure}[ht]
    \centering
    {\includegraphics[width=\linewidth]{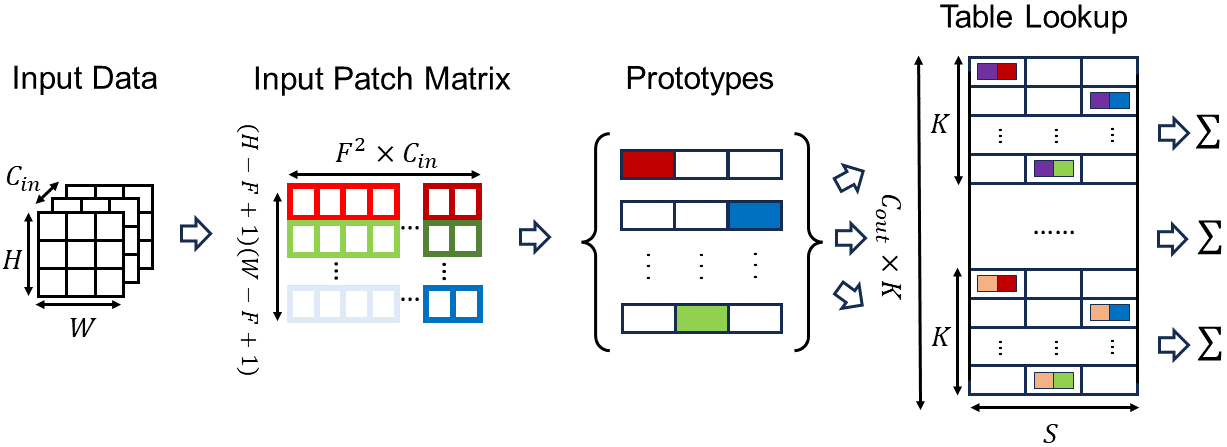}}
    \caption{Table lookup for convolutional operation inference.}
    \label{fig:conv_pq_query}
\end{figure}

\subsubsection{Table Lookups for Convolution Inference}
Figure~\ref{fig:conv_pq_query} shows the convolutional layer inference using the constructed table via lookup operations. Incoming query data (either an input image or a CNN layer input data) is reshaped to a patch matrix following the rule of im2col to a 2D matrix. Each row is split into $S$ subspaces. Then, for each subvector in a subspace, we find the corresponding prototype by running the trained clustering or hashing function. All the $HW\times S$ subvectors are independent and can run the prototype matching in parallel. Using the index of the matched prototypes, the precomputed dot product for each output channel can be directly acquired by looking up from the $C_{out}$ trained tables. The output lookup operations are also independent between output channels and can work in parallel. Finally, the subspace is aggregated through a simple sum operation, avoiding all MMs in convolution inference. 

\subsubsection{Linear Operation via Table Lookups}

\begin{figure}[ht]
    \centering
    {\includegraphics[width=\linewidth]{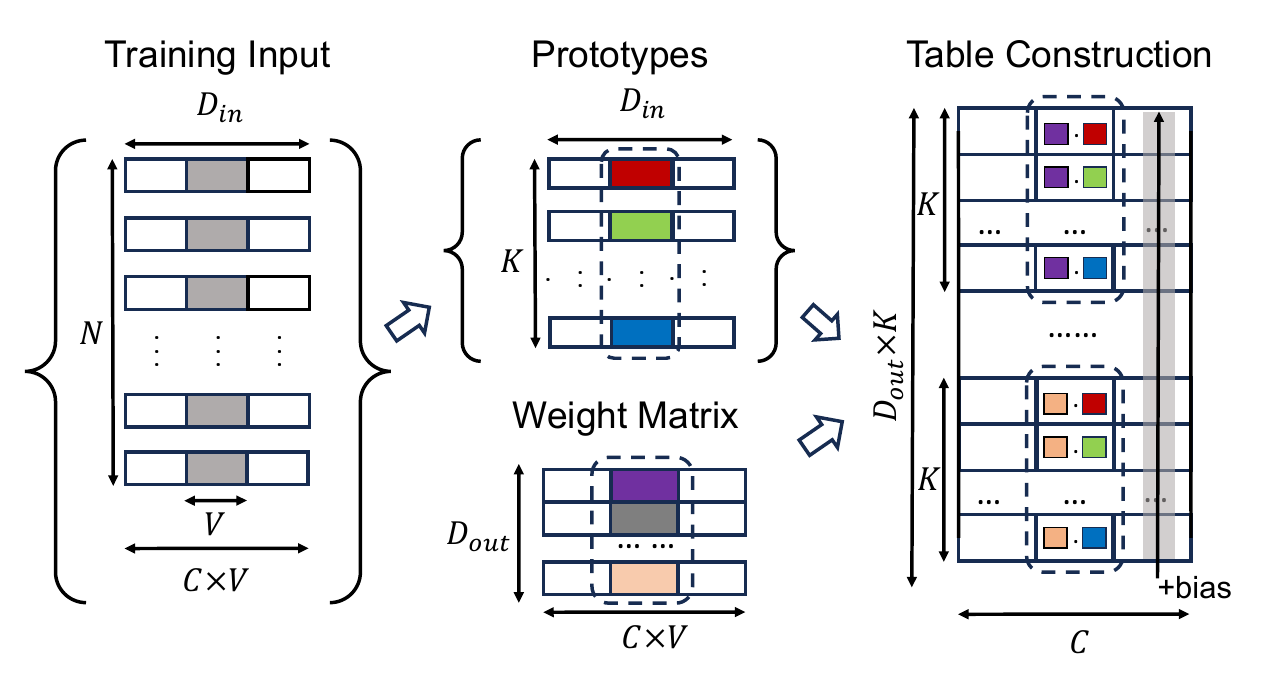}}
    \caption{Table construction for linear operation.}
    \label{fig:linear_pq_train}
\end{figure}

\begin{figure}[ht]
    \centering
    {\includegraphics[width=0.8\linewidth]{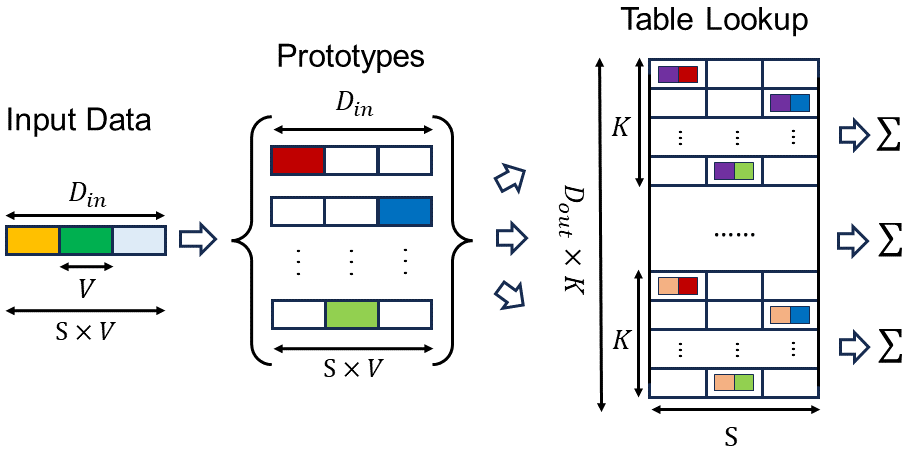}}
    \caption{Table lookup for for linear layer inference.}
    \label{fig:linear_pq_query}
\end{figure}

A linear operation is commonly used as the final classifier at the end of a CNN model. It transforms input $\mathbf{x}$ into an output $\mathbf{y}$ through a linear transformation, defined by the equation:
\begin{equation}
\mathbf{y} = \mathbf{Wx} + \mathbf{b}
\end{equation}
where $\mathbf{W}$ is the layer's weight matrix and $\mathbf{b}$ is the layer's biases. Figure~\ref{fig:linear_pq_train} shows the process of constructing tables for a linear layer. For a training set of $N$ inputs with input dimension $D_{in}$, we divide $D_{in}$ into $S$ subspaces, each with $K$ prototypes. The linear layer weights are split similarly. We create a table with $D_{out}$ sub-tables of $K\times S$ entries by storing dot product results between weight and prototype subvectors. To include the bias from the linear layer, we add it to a table column, ensuring its inclusion during the final aggregation in inference, as depicted in Figure~\ref{fig:linear_pq_query}. The linear layer inference then relies solely on prototype matching via hashing and table lookups, similar to table-based convolution.


\subsubsection{Folding Batch Normalization}
\label{sec:approach-bn}
{\color{black}
Batch normalization is widely used to speed up training and provide regularization for deep CNNs~\cite{ioffe2015batch}. While batch normalization helps convergence, it requires an inference calculation after each convolution.
To further reduce operations in a CNN model, we fold batch normalization operations into the convolution process, merging these two layers by transforming the weights and bias using the following equations:
}
\begin{equation}
\begin{gathered}
\label{eq:bn}
\mathbf{W^{\prime}}=\mathbf{W} \frac{\gamma}{\sqrt{\sigma^2+\epsilon}} \\
\mathbf{b}^{\prime}=\frac{\gamma}{\sqrt{\sigma^2+\epsilon}}(\mathbf{b}-\mu)+\beta
\end{gathered}
\end{equation}
where $\gamma$ and $\beta$ are trained parameters, $\sigma^2$ is the variance, $\mu$ is the mean, and $\epsilon$ is a small constant.

\subsubsection{Activation Function}
The most commonly used non-linear activation function in CNNs is ReLU~\cite{agarap2018deep}. This step is combined with aggregation after table lookups, as shown in Equation~\ref{eq:act}.
\begin{equation}
\label{eq:act}
    f_{act}\left(\mathbf{a}, \mathbf{b}\right) = \operatorname{max}\{0,\sum_s h^{s}\left(\mathbf{b}\right)_k\}, k = g^{s}\left(\mathbf{a}\right)
\end{equation}

\begin{algorithm}[t]
  \caption{Priority Masking}
  \label{alg:tabularize}
  \begin{algorithmic}[1]
    \State \textbf{Input:} Trained $N$-layer CNN model $\mathcal{M}$
    \State \textbf{Input:} Trained table-based approximation model $\mathcal{T}$ 
    \State \textbf{Input:} Training input data $\mathcal{D}$
    \State \textbf{Input:} Priority masking rate $M$
    \State \textbf{Initialize:} Layer similarity list $Sim$ of size $N$
    \State \textbf{Initialize:} Similarity difference list $SimDrop$ of size $N-1$
    \State \textbf{Initialize:} Priority list $Priority$ of size $N-1$
    \State \textbf{Initialize:} Mask list $Mask$ of size $N$, initialized as 1
    \For{$i$ \textbf{in} $0$ to $N-1$} \Comment{Get layer-wise similarity drop}
        \State $\mathbf{y_i} \leftarrow \mathcal{M}[0:i](\mathcal{D})$ 
        \State $\mathbf{\hat{y}_i} \leftarrow \mathcal{T}[0:i](\mathcal{D})$ 
        \State $Sim[i] \leftarrow  S_C(\mathbf{y_i},\mathbf{\hat{y}_i})$ 
        \If {$i>0$}: 
            \State $SimDrop[i]\leftarrow \left(Sim[i] - Sim[i-1], i\right)$
        \EndIf
    \EndFor
    \State $Priority \leftarrow$ Sorted($SimDrop$, Reverse=True)
    \For{$i$ \textbf{in} $0$ to $\lfloor M (N-1) \rceil$}   \Comment{Mask list update}
        \State j = $Priority$[i][1]
        \State $Mask$[j] = 0
    \EndFor 
    \For{$i$ \textbf{in} $0$ to $N-1$} \Comment{Retrain using mask list}
        \If{$Mask[i]=0 $}
        \State $\mathcal{T}[i] \leftarrow \mathcal{M}[i]$
        \Else
        \State $\mathcal{T}[i] \leftarrow \text{Retrain}\left(\mathcal{T}[i]\right)$
        \EndIf
    \EndFor
  \end{algorithmic} 
\end{algorithm}

\begin{table*}[t]
  \caption{Complexity analysis of TabConv and state-of-the-art acceleration methods.}
  \label{tab:baseline-flops}
  \begin{center}
    \begin{small}
    \begin{tabular}{lcccc}
    \toprule
     & \multicolumn{2}{c}{Arithmetic Operations (MFLOPs)}  & \multicolumn{2}{c}{Storage Cost (MBytes)} \\
    \cmidrule(lr){2-3}
    \cmidrule(lr){4-5}
    Approach&Expression$^\ast$& Example$^\dag$& Expression & Example\\

    \midrule
    im2col~\cite{chetlur2014cudnn}  &$C_{out}H'W'(2F^{2}C_{in}-1)$ & 5.105 &$C_{in}C_{out}F^{2}d$ & 0.036 \\
    kn2row~\cite{vasudevan2017parallel} & $C_{out}H'W'(2F^{2}C_{in}-1)$ & 5.105 &$C_{in}C_{out}F^{2}d$ & 0.036 \\
    Toeplitz~\cite{chen2020survey} & $C_{out}H'W'(2F^{2}C_{in}-1)$ & 5.105 &$C_{in}C_{out}F^{2}d$ & 0.036  \\
    FFT~\cite{chi2020fast} & $\beta C_{in}C_{out}H'W'+\alpha^{2}C_{out}H'W'\left(\frac{1}{2}C_{in}+\frac{13}{16}C_{out}\right)$ & 0.272 &$\beta C_{in}C_{out}F^{2}d+\alpha^{2}C_{out}d\left(\frac{1}{2}C_{in}+\frac{3}{2}C_{out}\right)$ & 0.032\\
    TabConv  & $H'W'\bigl[S\log_2 (K) + C_{out}\log_2 (S)\bigr]$ & 0.081& $H'W'S\log_2 (K)+C_{out}SKd$ &  16.02 \\
  \bottomrule
    \multicolumn{5}{l}{\small $\ast$ $H':= \frac{H-F+2P}{T}+1, W':= \frac{W-F+2P}{T}+1, \beta:=1-\alpha^2$}\\
    \multicolumn{5}{l}{\small $\dag$ Example based on the first convolution in ResNet-18: $H=32, W=32, F=7, P=3, T=2, C_{in}=3, C_{out}=64, \alpha =0.5, S = 8, K = 8192, d=4$}\\
\end{tabular}
\end{small}
\end{center}
\end{table*}

{\color{black}
\subsection{Priority Masking}}
\label{sec:masking}

When mapping more CNN layers to table lookups, the approximation error accumulates, and the prediction performance drops. To address this, we propose a novel priority masking strategy to retain exact operations crucial for maintaining model performance.

We introduce a priority masking rate $M\in[0,1]$ to adjust the masking, defined as the rate between the number of layers to be masked (not mapped to table) and the total number of model layers.

{\color{black}
Given a CNN model $\mathcal{M}$, its table-based approximation $\mathcal{T}$, and a priority masking rate $M$, we calculate the cosine similarity $S_C$ of each layer's output to compute the similarity drop between consecutive layers (lines 9-16). We sort the similarity drop list in descending order, giving layers with the most significant drops in $S_C$ the highest priority for masking (line 17). Based on this priority list, we update the mask list to 0 for layer indices holding the $\lfloor M(N-1)\rceil$ largest drops in similarity (lines 18-21). We then retrain the table-based model based on the mask list. Retraining the tables is necessary since the new architecture alters the input distribution to layers, affecting prototype matching and potentially amplifying existing input errors.
}

This selective masking ensures critical features are preserved, maintaining the accuracy of the model's inference while still benefiting from the efficiency of table lookups where appropriate.

\subsection{Complexity Analysis}

We analyze the computational complexity of our approach by examining arithmetic operations--an established indicator of inference latency \cite{chi2020fast, lavin2016fast}--and storage costs.

\subsubsection{Arithmetic Operations}
Arithmetic operations result from the following two processes: vector encoding $g$ to get table indices and aggregation $f$ after the table lookup to output results. For a convolution of an input with $C_{in}$ channels of size $H\times W$, stride $S$, and padding $P$ with $C_{out}$ kernels of size $F\times F$, it follows that $g$ results in $H'W'S\log_2 (K)$ operations and $f$ results in $H'W'C_{out}log_2 (S)$ operations, where $H'=\frac{H-F+2P}{T}+1$ and $W'=\frac{W-F+2P}{T}+1$. The total number of arithmetic operations (FLOPs) is then:
\begin{equation}
    \label{eq:arithmetic_operations}
    \begin{aligned}
        H'W'\bigl[S\log_2 (K) + C_{out}\log_2 (S)\bigr]
    \end{aligned}
\end{equation}

\subsubsection{Storage Cost}

Storage costs consist of the vector encoding results and table entries, where one index of an encoded prototype incurs a cost of $\log_2 (K)$ bytes and we denote the data byte-length of a precomputed entry as $d$. The actual prototypes need not be stored, as we use the encoded indices to look up table results directly. Considering the same convolution setup and parameters as in Section 4.5.1, there are $H'W'$ prototype indices and $C_{out}SK$ total table entries. Thus, the resulting storage in bytes is:
\begin{equation}
    \label{eq:storage_cost}
    \begin{aligned}
        H'W'S\log_2 (K)+C_{out}SKd
    \end{aligned}
\end{equation}


\subsubsection{Comparison with State-of-the-Art}
We compare our approach to various widely used, well-optimized algorithmic acceleration methods, each of which is outlined below.
\begin{itemize}
    \item \textbf{im2col}~\cite{chetlur2014cudnn}: Im2col, as seen in Section \ref{sec:background-im2col}, transforms convolution into matrix multiplication by creating patch matrices from the input image and kernels.
    { \color{black}
    \item \textbf{kn2row}~\cite{vasudevan2017parallel}: Kn2row turns each $F\times F$ convolution into $F^{2}$ $1\times 1$ convolutions by shifting the final feature map.
    \item \textbf{Toeplitz}~\cite{chen2020survey}: Toeplitz convolutions convert the input image to a Toeplitz matrix by unrolling kernel-sized patches. 
    \item \textbf{Fast Fourier Transform (FFT)}~\cite{chi2020fast}: FFT convolutions efficiently compute large kernel convolutions using Fourier transforms to avoid direct spatial convolutions.
    }
\end{itemize}
Table \ref{tab:baseline-flops} describes the number of FLOPs and parameter counts for these methods alongside our TabConv implementation. $C_{in}$ and $C_{out}$ represent the number of input and output channels of an arbitrary image, respectively, while $H$ and $W$ signify its dimensions, $F$ denotes kernel size, $T$ stride, $P$ padding, and $d$ the data byte length ($H'$ and $W'$ are the output dimensions, calculated using the expressions outlined in $*$ in the Table legend). Additionally, $\alpha \in[0, 1]$ in the FFT row is a parameter that varies by layer and kernel size \cite{chi2020fast}, and $S$ and $K$ in the TabConv row are the number of subspaces and prototypes per subspace, respectively.

We provide example FLOPs and storage calculations for implementation in the first convolution of ResNet-18, with parameters as stated in $\dag$. Our work results in a $98.4$\% decrease in MFLOPs compared to im2col, kn2row, and Toeplitz, and a $70.2$\% decrease in MFLOPs compared to FFT. The configuration $S=8$, $K=8192$ in this case incurs large storage costs. Using $S=2$, $K=2048$ would increasingly reduce MFLOPs while requiring a lesser $29.3$x storage compared to other approaches. Though storage cost is non-negligible, the number of arithmetic operations is largely reduced.
\\


\section{Experiments}

\subsection{Experimental Setup}

\subsubsection{Models} 
We apply our approach, TabConv, to three distinct CNN models for evaluation, including ResNet-18, ResNet-34, and NetworkInNetwork (NIN). The models complexity of accuracy performance on a variety of datasets are shown in Table~\ref{tab:models}. The models, selected for their varied architectural features, are as follows:
\begin{itemize}
    \item \textbf{ResNet-18}~\cite{he2016deep}: ResNet-18 is a part of the Residual Network family with 18 layers organized into blocks, allowing for apt training and better performance by using skip connections. 
    \item \textbf{ResNet-34}~\cite{he2016deep}: ResNet-34 extends ResNet-18 by increasing its depth to 34 layers. It leverages extended residual blocks and skip connections to best capture complex features.
    \item \textbf{Network In Network (NIN)}~\cite{lin2014network}: NIN integrates micro networks as multilayer perceptrons into convolutional layers. We implement a 9 layer NIN for evaluation.
\end{itemize}

\begin{table}[h]
  \caption{Complexity and accuracy of models we implemented.}
  \label{tab:models}
  \begin{center}
    \begin{small}
    \begin{tabular}{lccccc}
    \toprule
      & \multicolumn{2}{c}{Complexity}  & \multicolumn{3}{c}{Accuracy} \\
    \cmidrule(lr){2-3}
    \cmidrule(lr){4-6}
    Model& MFLOPs & Size (MB)& C10 & C100 & MN\\
    \midrule
    ResNet-18~\cite{he2016deep}  & 37.67 & 47.76 & 0.844 & 0.493 & 0.983 \\
    ResNet-34~\cite{he2016deep}  & 75.49 & 87.19  & 0.821 & 0.515 & 0.986 \\
    NIN~\cite{lin2014network} & 223.90 & 3.78 & 0.851 & 0.319 & 0.879 \\
    \bottomrule
    \end{tabular}
    \end{small}
    \end{center}
    \end{table}



\subsubsection{Datasets}


\ourwork~is assessed on three datasets, consisting of varying image complexities and class diversities, as follows:
\begin{itemize}
    \item \textbf{CIFAR-10~(C10)}~\cite{krizhevskycifar10} with 60K images and 10 classes.
    \item \textbf{CIFAR-100~(C100)}~\cite{krizhevskycifar100} with 60K images and 100 classes.
    \item \textbf{MNIST (MN)}~\cite{deng2012mnist} with 70K images and 10 classes.
\end{itemize}

\subsubsection{Metrics}

To comprehensively evaluate our approach against standard CNN performance, we consider the following metrics:


\begin{itemize}
    \item \textbf{Accuracy:} determined by the percentage of test images correctly predicted as their true class, measuring the model's ability in prediction.
    \item \textbf{Arithmetic operations in FLOPs:} the number of floating-point operations (FLOPs) needed for a single inference, measures a model's computational complexity.
    
    \item \textbf{Storage cost in Bytes:} the number of bytes required to store the model, assessing the memory footprint of the model.
\end{itemize}

\subsection{Evaluation of Table Lookup Mapping}
We thoroughly investigate the performance of mapping CNNs to table lookups by tuning the following configurations: number of subspaces $S$, number of prototypes per subspace $K$, and priority masking rate $M$. Based on a case configuration $S=8, K=8192$ (8\text{K}), and $M=0.4$, we explore the design space of each dimension through variable control, as shown in Table~\ref{tab:vary-s}, Table~\ref{tab:vary-k}, and Table~\ref{tab:vary-m}.

\begin{table}[h!]
  \caption{Accuracy when varying the number of subspaces $S$.}
  \setlength{\tabcolsep}{2.6pt} 
  \label{tab:vary-s}
    \begin{small}
    \begin{tabular}{ccccccccccccc}
    \toprule
    && & \multicolumn{3}{c}{ResNet18} & \multicolumn{3}{c}{ResNet34} & \multicolumn{3}{c}{NIN}  \\
    \cmidrule(lr){4-6}
    \cmidrule(lr){7-9}
    \cmidrule(lr){10-12}
    $S$ & $K$ & $M$ & C10 & C100 & MN  & C10 & C100 & MN& C10 & C100 & MN  \\
    \midrule
    1 &8K&0.4& 0.540 & 0.134 & 0.970 & 0.586 & 0.108 & 0.972 & 0.164 & 0.032 & 0.646 \\
    2 &8K&0.4& 0.542 & 0.116 & 0.980 & 0.636 & 0.130 & 0.974 & 0.274 & 0.038 & 0.716 \\
    4 &8K&0.4& 0.634 & 0.162 & 0.974 & 0.698 & 0.126 & 0.986 & 0.298 & 0.048 & 0.804  \\
    8 &8K&0.4& 0.696 & 0.172 & 0.982 & 0.728 & 0.164 & 0.986 & 0.472 & 0.064 & 0.812 \\
    16&8K&0.4& 0.766 & 0.218 & 0.984 & 0.780 & 0.204 & 0.986 & 0.612 & 0.118 & 0.822 \\
    \bottomrule
  \end{tabular}
    \end{small}
\end{table}

\begin{table}[h!]
  \caption{Accuracy when varying the number of prototypes $K$ per subspace.}
  \label{tab:vary-k}
  \setlength{\tabcolsep}{2.6pt} 
    \begin{small}
    \begin{tabular}{ccccccccccccc}
    \toprule
    &&  & \multicolumn{3}{c}{ResNet-18} & \multicolumn{3}{c}{ResNet-34} & \multicolumn{3}{c}{NIN}  \\
    \cmidrule(lr){4-6}
    \cmidrule(lr){7-9}
    \cmidrule(lr){10-12}
    $S$ & $K$ & $M$ & C10 & C100 & MN  & C10 & C100 & MN& C10 & C100 & MN  \\
    \midrule
    8 &1K&0.4& 0.684 & 0.162 & 0.978 & 0.736 & 0.132 & 0.978 & 0.336 & 0.040 & 0.726 \\
    8 &2K&0.4& 0.686 & 0.186 & 0.980 & 0.710 & 0.142 & 0.978 & 0.366 & 0.060 & 0.774 \\
    8 &4K&0.4& 0.722 & 0.180 & 0.982 & 0.724 & 0.154 & 0.980 & 0.422 & 0.064 & 0.806 \\
    8 &8K&0.4& 0.696 & 0.172 & 0.982 & 0.728 & 0.164 & 0.986 & 0.472 & 0.064 & 0.812 \\
    8&16K&0.4& 0.684 & 0.136 & 0.984 & 0.718 & 0.110 & 0.978 & 0.512 & 0.060 & 0.816 \\
    \bottomrule
  \end{tabular}
    \end{small}
\end{table}

\begin{table}[h!]
  \caption{Accuracy when varying the priority masking rate $M$.}
  \label{tab:vary-m}
  \setlength{\tabcolsep}{2.6pt} 
    \begin{small}
    \begin{tabular}{ccccccccccccc}
    \toprule
    &&  & \multicolumn{3}{c}{ResNet-18} & \multicolumn{3}{c}{ResNet-34} & \multicolumn{3}{c}{NIN}  \\
    \cmidrule(lr){4-6}
    \cmidrule(lr){7-9}
    \cmidrule(lr){10-12}
    $S$ & $K$ & $M$ & C10 & C100 & MN  & C10 & C100 & MN& C10 & C100 & MN  \\
    \midrule
    8 &8K&0 & 0.248 & 0.068 & 0.964 & 0.284 & 0.058 & 0.964 & 0.276 & 0.042 & 0.818 \\
    8 &8K&0.2& 0.590 & 0.130 & 0.976 & 0.416 & 0.096 & 0.986 & 0.372 & 0.038 & 0.814 \\
    8 &8K&0.4& 0.696 & 0.172 & 0.982 & 0.728 & 0.164 & 0.986 & 0.472 & 0.064 & 0.812 \\
    8 &8K&0.6& 0.840 & 0.254 & 0.988 & 0.806 & 0.220 & 0.988 & 0.474 & 0.186 & 0.838 \\
    8 &8K&0.8& 0.822 & 0.492 & 0.984 & 0.820 & 0.408 & 0.986 & 0.500 & 0.228 & 0.852 \\
    \bottomrule
  \end{tabular}
    \end{small}
\end{table}

In Table \ref{tab:vary-s}, we examine how changing the number of subspaces ($S$) from 1 to 16 affects the accuracy of table-based model inferences. For ResNet-18, the accuracy drop on CIFAR-10, CIFAR-100, and MNIST is 0.226, 0.084, and 0.014 respectively. Similarly, for ResNet-34 and NIN across these datasets, we observe accuracy drops of 0.194, 0.096, 0.014, and 0.448, 0.086, 0.176 respectively with decreasing $S$. The results indicate that more subspaces enhance the approximation capability of table-based models.

In Table \ref{tab:vary-k}, we explore the influence of varying the number of prototypes per subspace ($K$) from 1K to 16K on model accuracy. Notably, in ResNet-18 on CIFAR-10, the performance drops by 0.038 when $K$ changes from 4K to either 1K or 16K, suggesting an $K$ for certain cases. While increasing $K$ to 16K maximizes performance in specific scenarios, such as ResNet-18 for MNIST and NIN for CIFAR-10 and MNIST, too many prototypes can sometimes result in lower accuracy. Thus, carefully calibrating $K$ is crucial, especially when operations are more straightforward to approximate. Despite this, increasing $K$ provides higher accuracy in most cases even when the best accuracy is not from the maximum value of $K$.

Table \ref{tab:vary-m} shows the impact of varying the priority masking rate ($M$) over 0, 0.2, 0.4, 0.6, and 0.8's effect on the accuracy of~\ourwork~inference. When decreasing $M$ from 0.8 to 0, ResNet-18 on CIFAR-10, CIFAR-100, and MNIST has an accuracy drop of 0.574, 0.424, 0.020 respectively. For ResNet-34 on CIFAR-10, CIFAR-100, and MNIST, table-based models undergo an accuracy drop of 0.536, 0.350, and 0.022 respectively. For NIN on CIFAR-10, CIFAR-100, and MNIST, table-based models undergo an accuracy drop of 0.224, 0.186, and 0.034 respectively. In particular, the accuracy drops for the MNIST dataset are low and tell that when models are able to achieve certain performance on specific datasets, increasing $M$ does not help as much as it would on harder to approximate data.

\subsection{Evaluation of Priority Masking}



In evaluating the priority masking technique, we methodically replace 20\%, 40\%, 60\%, and 80\% of matrix multiplication operations in the forward pass with exact neural network counterparts based on cosine similarity, prioritizing layers with significant similarity reductions ($M = 0.2, 0.4,0.6,$ and $0.8$). This strategy helps us to measure the effects of substituting matrix operations with exact calculations on a per-layer basis, highlighting key trade-offs between accuracy, computational complexity, and storage requirements.

\begin{figure}[t]
    \centering
    \includegraphics[width=\linewidth]{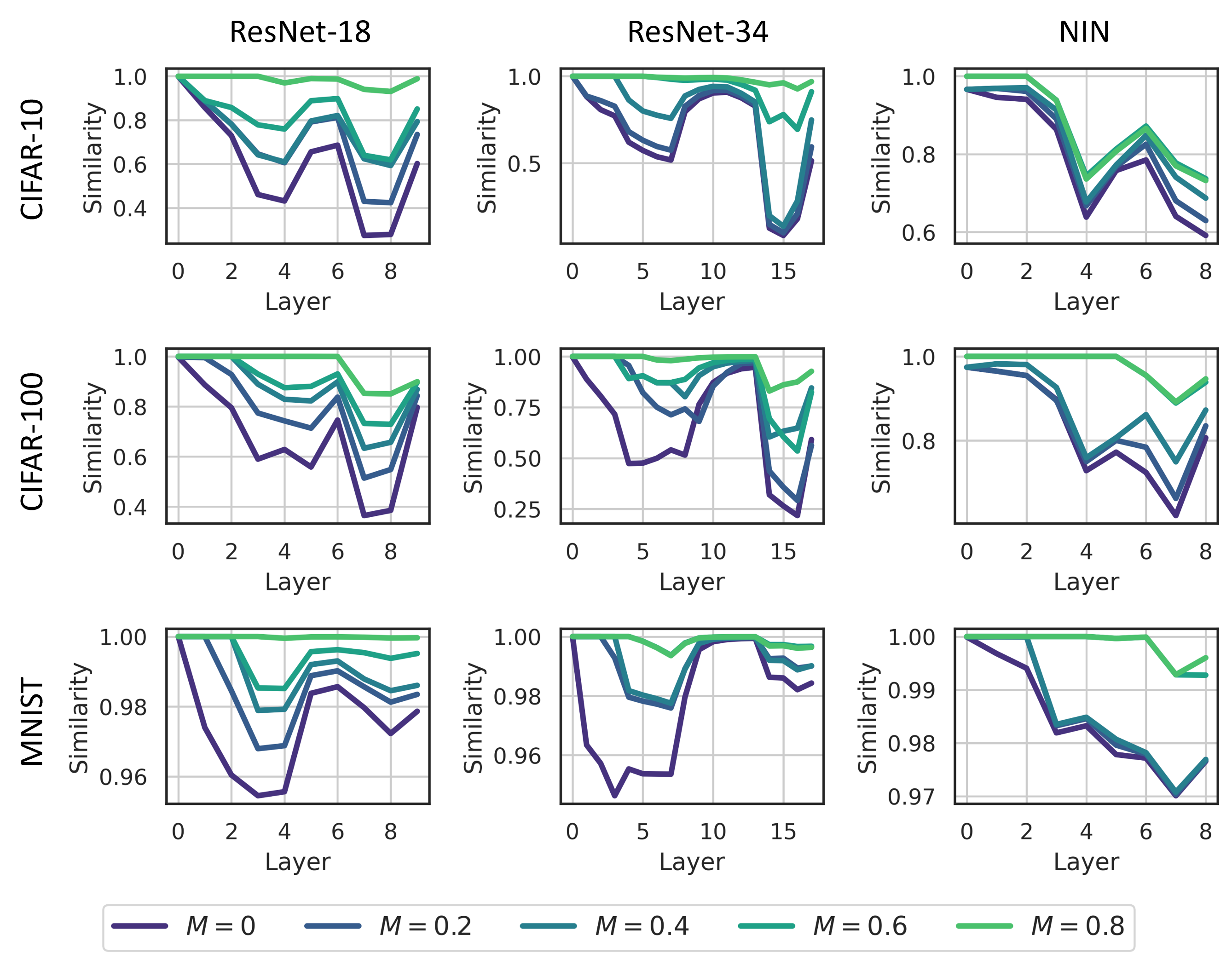}
    \caption{Layer-wise cosine similarity ($S_C$) when varying $M$.}
    \label{fig:cosine-sim}
\end{figure}

In Figure \ref{fig:cosine-sim}, we present the curve of $S_C$ for all models and all datasets with $S=8$ subspaces and $K=8K$ prototypes.
For ResNet-18 on CIFAR-10 dataset, the full table average layer-wise $S_C$ drop is 0.402. However, this decreases significantly to 0.289, 0.245, 0.182, and 0.019 as the masking rate ($M$) is increased to $0.2, 0.4,0.6,$ and $0.8$, respectively. Furthermore, $S_C$ in the last layer increases from 0.602 for the full table to 0.735, 0.793, 0.851, and 0.989 for varying values of $M$.
This trend holds for all models over all datasets but not to the same severity. Since degree of approximation's closeness impacts the final accuracy, we want to minimize layer-wise drops in $S_C$ as well as maximize $S_C$ in last layer. In the case of NIN, applying more aggressive masking percentages does not improve accuracy nor cosine similarity to the same degree. We reason this is due to varying model architectures since NIN does not contain a linear layer, making final layer approximations more difficult.

\subsection{Evaluation of Complexity}

\begin{figure}[t]
    \centering
    \includegraphics[width=\linewidth]{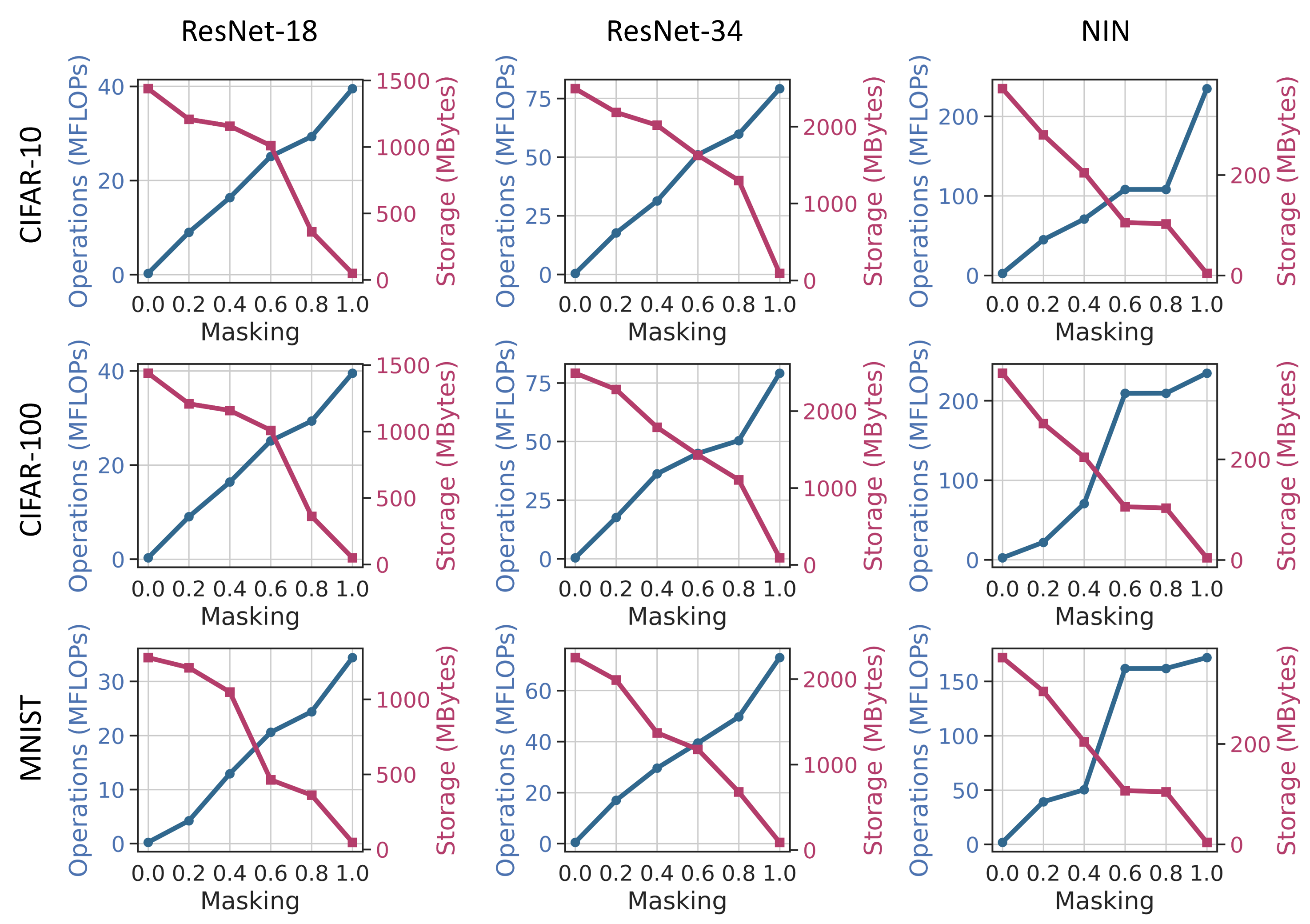}
    \caption{Operations and storage costs when varying $M$.
    }
    \label{fig:costs}
\end{figure}

We comprehensively evaluate table-based models' complexity and tradeoffs, noting our work is a time-space tradeoff approach. Figure \ref{fig:costs} shows the trade off between storage and number of operations for table-based models under varying values of $M$. For example in CIFAR-10, ResNet-18 shows a FLOPs decrease from 39.5 million to as low as 260,000 in the process of converting the full CNN to tables. For ResNet-34, the reduction is from 79.1 million to around 388,000 FLOPs. The NIN model shows a significant drop from 234.8 million to about 2.5 million FLOPs. In terms of storage, ResNet-18's memory footprint goes from 11.7 MB to 343.0 MB when converting all operations to table lookups. ResNet-34's memory footprint goes from 21.8 MB to 594.5 MB when converting all operations to table lookups. NIN's memory footprint goes from 997.0 KB to 88.6 MB when converting all operations to table lookups.

\subsection{Evaluation of Overall Performance}


We introduce three metrics to evaluate the relative performance against standard CNNs: $R_A$, $R_F$, and $R_S$ where $R_A$ measures the accuracy ratio, $R_F$ measures the ratio of total operations, and $R_S$ measures the storage cost ratio between \ourwork~and CNNs.


In Table~\ref{tab:key_res}, we summarize the selection of priority masking rate $M$ and~\ourwork~overall performance given the lower bounds of $R_A$ at 0.9, 0.8, and 0.7. 
For a bound $R_A \geq 0.9$, \ourwork~retains at least 93\% of the original model's accuracy ($R_A\geq0.93$), achieving a reduction in FLOPs of 36.5\%, 25.8\%, and 99.4\% for ResNet-18 on CIFAR-10, CIFAR-100, and MNIST, respectively. 
It also reduces FLOPs by 35.6\% and 99.3\% for ResNet-34 on CIFAR-10 and MNIST, respectively, and achieves a 98.9\% reduction in FLOPs for NIN on MNIST.
When attaining over 80\% of initial CNN accuracy, \ourwork~is able to reduce 58.6\% and 60.5\% of total operations during inference for ResNet-18 and ResNet-34 on CIFAR-10 respectively. 
When attaining over 70\% of CNN accuracy, \ourwork~is able to reduce 77.2\% of FLOPs for ResNet-18 on CIFAR-10, reduce 36.3\% FLOPs for ResNet-34 on CIFAR-100, and reduce 10.8\% FLOPs for NIN on CIFAR-100.


\ourwork-based models exhibit a space-time tradeoff, necessitating on average a 32.0$\times$, 33.4$\times$, and 29.8$\times$ increase in memory footprint to retain 90\%, 80\%, and 70\% accuracy, respectively, compared to CNN-based models. 
In general, our work effectively reduces the number of operations required for CNN inference while maintaining model accuracy, albeit with increased storage costs.


\begin{table}[t]
  \caption{TabConv masking selection and performance when maintaining thresholds of original CNN accuracy.}
    \setlength{\tabcolsep}{3pt} 
  \label{tab:key_res}
    \begin{small}
    \begin{tabular}{ccccccccccc}
    \toprule
    && \multicolumn{3}{c}{ResNet-18} & \multicolumn{3}{c}{ResNet-34} & \multicolumn{3}{c}{NIN} \\
    \cmidrule(lr){3-5}
    \cmidrule(lr){6-8}
    \cmidrule(lr){9-11}
    Bound   &  Res & C10& C100 & MN & C10 & C100 & MN& C10 & C100 & MN \\
        \hline
    &    $M~$ & 0.6 & 0.8 & 0 & 0.6 & - & 0 & - & - & 0 \\
    $0.9$   &    $R_A$ & 0.995 & 0.998 & 0.981 & 0.982 & - & 0.984 & - & - & 0.931 \\
    &    $R_F$ & 0.636 & 0.742 & 0.006 & 0.644 & - & 0.007 & - & - & 0.011 \\
    &    $R_S$ & 20.61 & 7.377 & 27.26 & 17.81 & - & 25.21 & - & - & 94.80 \\
        \hline
    &    $M~$ & 0.4 & 0.8 & 0 & 0.4 & - & 0 & - & - & 0 \\
    $0.8$   &    $R_A$ & 0.825 & 0.998 & 0.981 & 0.887 & - & 0.984 & - & - & 0.931 \\
    &    $R_F$ & 0.414 & 0.742 & 0.006 & 0.395 & - & 0.007 & - & - & 0.011 \\
    &    $R_S$ & 23.62 & 7.377 & 27.26 & 22.08 & - & 25.21 & - & - & 94.80 \\
        \hline
    &    $M~$ & 0.2 & 0.8 & 0.0 & 0.4 & 0.8 & 0 & - & 0.8 & 0 \\
    $0.7$   &    $R_A$ & 0.700 & 0.998 & 0.981 & 0.887 & 0.792 & 0.984 & - & 0.715 & 0.931 \\
    &    $R_F$ & 0.228 & 0.742 & 0.006 & 0.395 & 0.637 & 0.007 & - & 0.892 & 0.011 \\
    &    $R_S$ & 24.64 & 7.38 & 27.26 & 22.08 & 12.09 & 25.21 & - & 26.03 & 94.80  \\
    \bottomrule
  \end{tabular}
    \end{small}
\end{table}

\section{Conclusion}

We proposed \textit{\ourwork}, a novel CNN approximation based on converting key operations to table lookups. The key steps in our process include taking a trained CNN and converting all instances of convolution to matrix multiplication, mapping these to table lookups via product quantization, and employing a priority masking technique to identify which layers should be replaced by table approximation and which should retain their original form. Our \ourwork-based model significantly reduces arithmetic operations while maintaining performance. In future work, we plan to optimize the table-based approximation through developing automatic configuration tools for table structure, quantized table entries for compression, and better hashing functions for prototype matching.

\section*{Acknowledgment}

This work has been supported by the U.S. National Science Foundation (NSF) under grants CNS-2009057 and SaTC-2104264, as well as the DEVCOM Army Research Lab (ARL) under grant W911NF2220159.

\bibliographystyle{ACM-Reference-Format}
\bibliography{ref}

\end{document}